\documentclass{article}

\usepackage[nonatbib, final]{neurips_2024}

\usepackage[utf8]{inputenc} 
\usepackage[T1]{fontenc}    
\usepackage{hyperref}       
\usepackage{url}            
\usepackage{booktabs}       
\usepackage{amsfonts}       
\usepackage{nicefrac}       
\usepackage{microtype}      
\usepackage{xcolor}         
\usepackage{graphics}
\usepackage[pdftex]{graphicx}

\hypersetup{hidelinks} 
\usepackage{multirow}
\usepackage{mathtools,amssymb}
\usepackage{enumitem}
\usepackage[vlined,ruled,linesnumbered]{algorithm2e}
\usepackage{tabularx}
\usepackage[small,compact]{titlesec}
\newcolumntype{Y}{>{\centering\arraybackslash}X}

\title{TransVIP: Speech to Speech Translation System with Voice and Isochrony Preservation}

\author{%
  \small{Chenyang Le}$^{1,2}$\thanks{Work done during an internship at Microsoft Azure AI. nethermanpro@sjtu.edu.cn} 
  \And \small{Yao Qian}$^{2}$ \thanks{Correspondence: yaoqian@microsoft.com} 
  \And \small{Dongmei Wang}$^{2}$
  \And \small{Long Zhou}$^{2}$
  \And \small{Shujie Liu}$^{2}$
  \And \small{Xiaofei Wang}$^{2}$
  \And \small{Midia Yousefi}$^{2}$
  \And \small{Yanmin Qian}$^{1}$
  \And \small{Jinyu Li}$^{2}$
  \And \small{Sheng Zhao}$^{2}$
  \And \small{Michael Zeng}$^{2}$
  \And
  $^{1}$\texttt{Shanghai Jiao Tong University,  China} 
  \And
  $^{2}$\texttt{Microsoft, USA} 
}

\begin{document}

\maketitle

\begin{abstract}
There is a rising interest and trend in research towards directly translating speech from one language to another, known as end-to-end speech-to-speech translation. However, most end-to-end models struggle to outperform cascade models, i.e., a pipeline framework by concatenating speech recognition, machine translation, and text-to-speech models. The primary challenges stem from the inherent complexities involved in direct translation tasks and the scarcity of data.  In this study, we introduce a novel model framework TransVIP that leverages diverse datasets in a cascade fashion yet facilitates end-to-end inference through joint probability. Furthermore, we propose two separate encoders to preserve the speaker's voice characteristics and isochrony from the source speech during the translation process, making it highly suitable for scenarios such as video dubbing. Our experiments on the French-English language pair demonstrate that our model outperforms the current state-of-the-art speech-to-speech translation model.
\end{abstract}

\begin{figure}[h]
  \centering
  \includegraphics[width=0.95\linewidth]{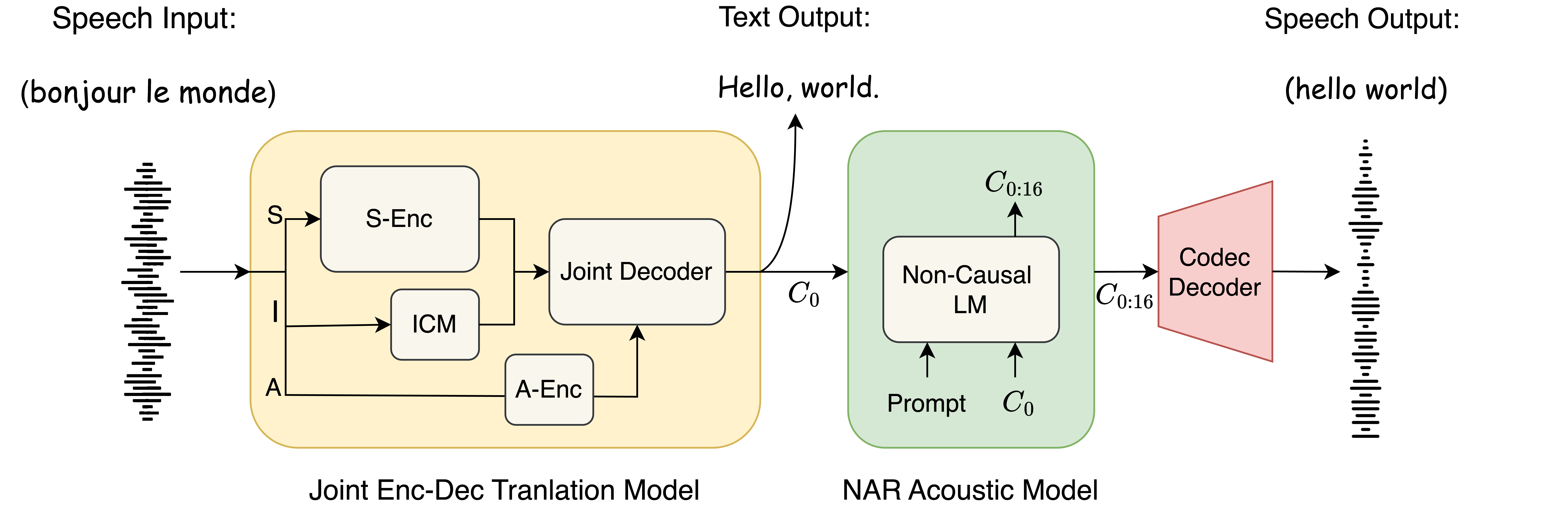}
  \caption{Overview of our speech-to-speech translation framework: 1) Joint encoder-decoder model for translating speech into the target text, and coarse-grained speech tokens, $C_0$; 2) Non-autoregressive acoustic model for acoustic details, $C_{0:16}$; 3) Codec model to convert discrete speech tokens back to the waveform. Abbreviation: S/A/I(Semantic/Acoustic/Isochrony Information), $C_0$/$C_{0:16}$(Codec layer 0/0-15), S/A-Enc(Semantic/Acoustic Encoder), ICM(Isochrony Control Module).}
  \label{fig:1}
\end{figure}

\section{Introduction}
In recent years, speech translation (ST) has shifted from loosely coupled cascaded systems to more integrated, and even end-to-end systems \cite{sperber2020speech}. Traditionally, cascaded systems comprised separate components for automatic speech recognition (ASR), machine translation (MT), and optionally, text-to-speech (TTS). Recent studies \cite{radford2023Robust, zhang2022speechut,zhang2023Google, le2023ComSL} have successfully integrated ASR and MT into a unified end-to-end speech-to-text translation (S2TT) system. Furthermore, the integration of TTS to form a speech-to-speech translation (S2ST) system has become an increasingly researched area.

The primary challenge in developing end-to-end S2ST systems lies in performance issues. Both S2TT and TTS involve high variability with multiple reasonable outputs. Simultaneously performing these tasks increases the complexity of learning exponentially. Additionally, there is a scarcity of end-to-end S2ST data. Most available data are weakly supervised, obtained through synthesis \cite{huang2022TranSpeech} or internet parsing \cite{communication2023SeamlessM4T}, further complicating the task. Another significant challenge is preserving speaker identity, as obtaining large-scale datasets with ground-truth paired speech spoken by the same speaker in two languages is nearly impossible. In practical applications like video dubbing, there is also a demand for controlling the length of the generated target speech, ensuring that it closely matches the length of the source speech. This capability of isochrony control, however, is absent in the majority of existing S2ST systems.

To address these issues, we propose TransVIP, a speech-to-speech \textbf{Trans}lation framework with \textbf{V}oice and \textbf{I}sochrony \textbf{P}reservation. TransVIP employs a consecutive generation approach, simplifying the complex S2ST task into two sequential tasks while maintaining an end-to-end framework. The generation of the target speech is conditioned not only on the semantic information, as in conventional S2ST models, but also on the isochrony and acoustic information derived from the source speech. 
The corresponding overview of TransVIP is demonstrated in Figure \ref{fig:1}. 
We evaluate the performance of TransVIP for French-English mutual translation using a subset of the CVSS-T test set \cite{jia-etal-2022-cvss}.
The results demonstrate that TransVIP outperforms the publicly available SOTA models such as a larger Seamless Expressive model. The generated audio samples are available at https://aka.ms/transvip. The training code and script are available at https://github.com/nethermanpro/transvip.
The main contributions of the paper can be summarized as follows:
\vspace*{-0.2\baselineskip} 
\begin{enumerate}[leftmargin=*]
    \setlength{\topmargin}{0pt}
    \setlength{\itemsep}{0.1em}
    \setlength{\parskip}{0pt}
    \setlength{\parsep}{0pt}
    \item We introduce a framework for speech-to-speech translation tasks that employ a consecutive generation with joint inference. This method efficiently utilizes a variety of datasets through multi-task learning to overcome the challenge of scarce paired data during the training phase, while preserving the end-to-end nature during inference. 

    \item We propose to disentangle various information required to learn in the training stage by employing separated encoders.  It can enhance the transfer of voice characteristics and isochrony temporal alignment from the source to the target speech in the translation process. Additionally, it facilitates the design of lightweight modules for more effective individual information learning. 

   \item We advance the SpeechToknizer \cite{zhang2023SpeechTokenizera} technology for multi-lingual tasks by distilling the semantic information from a large-scale self-supervised model to the latest high-performing codec model. This advancement allows us to employ a textless non-autoregressive model for learning fine codec code generation without text labels, which is impractical in the conventional codec-based speech generation methods such as VALL-E \cite{wang2023neural}.

   \item We propose a method that refines the decoding process by incorporating a sampling mechanism within the Layer Beam Search framework, thereby enhancing the efficiency and effectiveness of non-autoregressive model decoding.  
\end{enumerate}


\section{Related Works}
\paragraph{Speech Quantization}
The task of speech quantization is to transform continuous speech features into discrete tokens. The quantization module is usually trained by self-supervised learning (SSL) methods. The speech tokens can be divided into two categories: semantic tokens and acoustic tokens\cite{borsos2023AudioLM}. The semantic tokens\cite{baevski2020Wav2vec, hsu2021HuBERT, chung2021W2vBERT, cheng2023Mu, ao2022SpeechT5, chiu2022Selfsupervised} are usually learned by context prediction task. This kind of token is rich in context information and is suitable for downstream tasks like recognition. The acoustic tokens\cite{zeghidour2022SoundStream, valin2012definition, dietz2015Overview, defossez2023High, kumar2023HighFidelitya}, also called the neural codec, is usually trained by reconstruction task through a discrete information bottleneck. This codec is suitable for audio compression and audio generation. SpeechTokenizer \cite{zhang2023SpeechTokenizera} combines the semantic token and acoustic token by semantic distillation. The latest works propose attribute disentanglement of neural codec for better generation task support.  FAcodec\cite{ju2024NaturalSpeech} further disentangle codec into content, prosody, acoustic, and timbre through supervision and reversed gradient. 

\paragraph{Speech Language Model}
With the recent advance in speech quantization and the strong in-context learning capability of the language model, the speech-language model has shown great potential in speech generation. The GSLM family\cite{lakhotia2021Generativea, kharitonov2022TextFree, nguyen2023Generative} use semantic tokens to train language models on speech continuation. VoxtLM \cite{maiti2024VoxtLM} integrates text tokens with semantic tokens to perform speech recognition, synthesis, and continuation in a single causal language model. AudioLM\cite{borsos2023AudioLM} uses both semantic tokens and acoustic tokens for high-quality speech continuation.  VALL-E \cite{wang2023neural} extends this idea to zero-shot TTS and proposes auto-regressively generating the first layer and non-autoregressively the rest layers of the codec codes. Later work\cite{zhang2023speak, kharitonov2023Speak, huang2023MakeAVoice} have enhanced the functionality of zero-shot TTS systems, such as building cross-lingual synthesis and accurate alignment between text control signal and acoustic tokens. 

\paragraph{End-to-End S2ST}
An end-to-end approach has been an ultimate goal for S2ST research. Some previous works\cite{jia2022Translatotron, nachmani2024Translatotron} have tried a direct end-to-end approach for S2ST.  However, by far such methods still suffer great performance issues. A solution is to include semantic tokens(text, phoneme, or semantic speech tokens) as an intermediate result. Some recent works\cite{lee2022Direct, huang2022TranSpeech,wei2023joint, diwan2023Unitbased} applied speech-to-unit translation(S2UT) that translate speech input into semantic speech tokens and then generate speech through a unit vocoder like Hifi-GAN. And UnitY\cite{inaguma2023UnitY} use separate speech-to-text and text-to-unit models that can be optimized together. However traditional vocoder based on semantic tokens like Hifi-GAN cannot generate high-quality speech. SeamlessExpressive\cite{communication2023Seamless} addresses this issue using a well-designed PRETSSEL vocoder.  PolyVoice\cite{dong2023PolyVoice} and AudioPaLM\cite{rubenstein2023AudioPaLM} employees use two causal models to first translate speech into semantic tokens, then generate acoustic tokens based on semantic tokens, and then convert acoustic tokens to output speech. VioLA\cite{wang2023VioLA} and MSLM-S2ST\cite{pengMSLMS2STMultitaskSpeech2024} adopt a similar approach but use a single causal model in a multi-tasking way. However, these approaches perform several separate auto-regressive generations during the inference process. This, to a certain degree, contradicts the principle of end-to-end processing.

\begin{figure}
  \centering
  \includegraphics[width=1\linewidth]{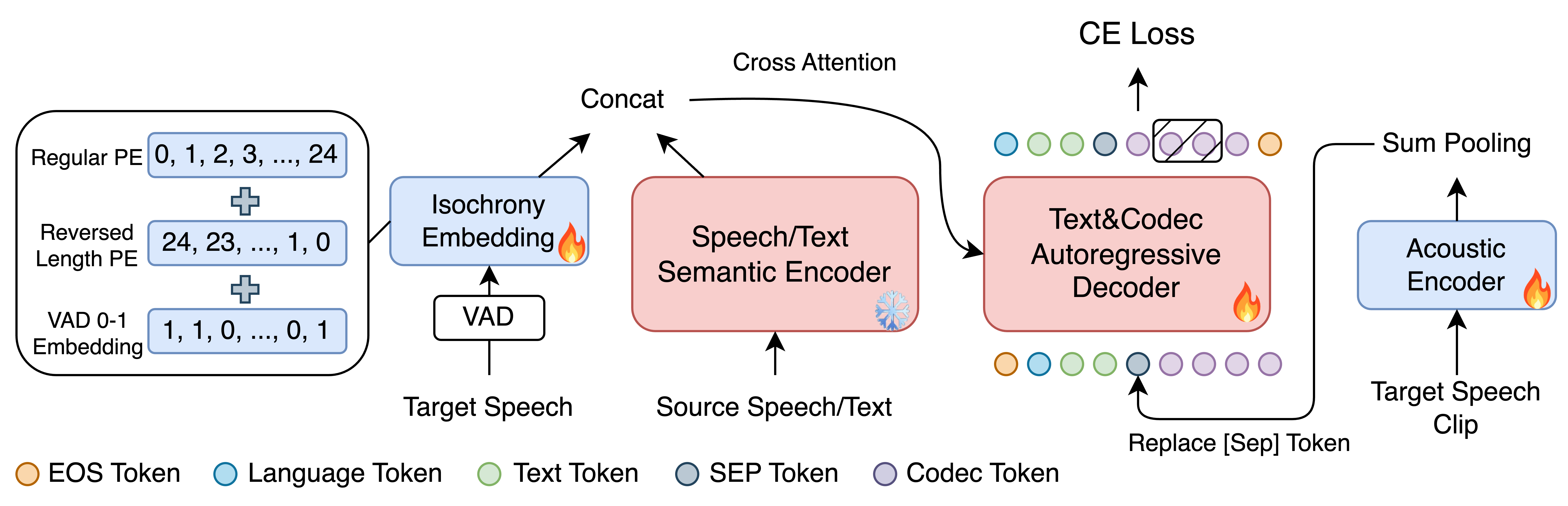}
  \caption{The illustration of the training framework of the Joint Enc-Dec Model. During the training, the losses from the target speech clip, i.e., a sub-part of the whole target speech, which serves as a prompt, are not aggregated when computing the Cross-Entropy (CE) loss. The corresponding codec labels are masked in the implementation. 
  The semantic encoder and the auto-regressive decoder are initialized by a SeamlessM4T X2T model \protect\footnotemark. The semantic encoder is frozen during training. In inference, all the target speech input are replaced by source speech input.}
  \label{fig:2}
\end{figure}

\section{TransVIP}
\label{TransVIP}
\subsection{Overview}\label{ar_section}
Inspired by the recent advance in codec-based zero-shot TTS such as VALL-E \cite{wang2023neural}, TransVIP contains three major parts: 1) A codec model for speech quantization and reconstruction, which is essential for auto-regressive speech generation (section \ref{codec_section}). 2) An auto-regressive joint translation model to translate the input speech into coarse-grained speech (Section \ref{ar_section}). 3) A textless non-autoregressive acoustic model to replenish acoustic details into the output speech (Section \ref{nar_section}). The overall framework for generating target speech is shown in Figure \ref{fig:1}.

\footnotetext{https://ai.meta.com/blog/seamless-m4t}\label{fn:m4t}

In this section, we will present a detailed introduction to our joint translation model, which is the key component of our translation system. The training procedure of the joint model is illustrated in Figure \ref{fig:2}. In section \ref{sec:other}, we will discuss the codec and NAR model.

\subsection{Consecutive Generation with Joint Inference}
\label{sec:consecutive}

Directly generating target translation speech $Y$ from input speech $X$ presents a considerable challenge. Therefore, the primary objective of this model is to produce a coarse-grained speech $Y'$ represented by a first-layer codec sequence. Instead of optimizing the direct probability distribution of $P(Y'|X)$, our optimizing target is the joint probability  $P(Y', T|X)$ to preserve semantic accuracy in the text outputs alone, as $P(T|X)$.
Hence, our approach focuses on modeling and maximizing the conditional expression:
\vspace*{-0.2\baselineskip} 
\begin{equation}
P(Y', T|X) =  P(Y'|X, T)P(T|X) 
\end{equation}
where $T$ represents the translated target text. This modeling is achieved through a consecutive generation framework and optimized using the beam search algorithm. Specifically, the decoder initially generates text tokens, followed by codec tokens. We employ a unique separation token to delineate the end of text tokens and the start of codec tokens. Consequently, the codec is conditioned on both the input speech and the generated text. In practical terms, exploring the entire space of $T$ is infeasible. In beam search generation, we optimize the following approximate form:
\vspace*{-0.2\baselineskip}
\begin{equation}
Y', T = \mathop{\arg\max}_{Y', T\in \mathcal{T}} P(Y'|X, T)P(T|X)
\end{equation}
where $\mathcal{T}$ represents a subset of $T$ with the highest speech-to-text translation probability selected by the beam search algorithm.

\subsection{Feature Disentanglement}
In the real world, an ideal training dataset for speech-to-speech generation is not readily available, where the source and target speeches are roughly equal in length and uttered by the same speaker. A real dataset comprises paired speech, either spoken by different speakers or synthesized.  Thus, we cannot guarantee voice and duration preservation with these corpora. 

To address this issue, we decouple the input feature into three parts: semantic information($S$), acoustic information($A$), and isochrony information($I$). Formally, $X=(S, A, I)$.  During training the inputs to the model are $S$ from source speech, and $(A, I)$ information from reference target speech. While during inference, $(A, I)$ are derived from source speech.  In this way the input feature can always match the output speech in terms of voice characterizes $A$ and isochrony $I$.

\paragraph{Semantic Information $S$} $S$ is extracted using a pre-trained speech encoder from an encoder-decoder speech/text-to-text translation model, designed to provide minimal acoustic or speaker information. We also include the text encoder of the pre-trained model to enable text input and perform a knowledge distillation loss on the text output part.  We freeze two pre-trained encoders during the training process.

\paragraph{Acoustic information $A$} 
The acoustic information that we extract relates to the distinctive acoustic characteristics of a speaker’s voice. We assume that the translated text $T$ is independent of $A$.
The relationship can be modeled as in Equation \ref{eqa:3}. We also take advantage of the one-direction mask in the auto-regressive decoder to model such a relation explicitly. 
\vspace*{-0.2\baselineskip} 
\begin{equation}
\label{eqa:3}
\begin{split}
    P(Y'|X, T)P(T|X) &= P(Y'|S, A, I, T)P(T|S, A, I) \\
                                &= P(Y'|S, A, I, T)P(T|S, I)
\end{split}
\end{equation}

We use an acoustic encoder, comprising transformer blocks that learn from scratch, to extract the acoustic information. This information is sum pooled along the time dimension into a single embedding, which then replaces the embedding of a separation token at the input of the auto-regressive decoder. This acoustic embedding will not affect text generation due to the causal property.  It is compatible with both gradients passing during training and beam search during inference. In training, the input to the acoustic encoder is a speech prompt, i.e., a sub-part of the whole target speech. We zero out the loss in the same sub-part of the label to prevent information leakage. This combined with the sum pooling forms an information bottleneck to prevent complex semantic information from passing, allowing the encoder to learn meaningful acoustic information for voice preservation. Like Classifier-free Guidance, this replacement operation is not performed with a certain probability(in our experiment $p=0.5$) during training.

\paragraph{Isochrony information $I$} We define a frame length, in our case 160ms, to quantify the duration of speech. Our Isochrony control module contains three embeddings: a regular position embedding, a reversed position embedding, and a voice activity detection (VAD) 0-1 embedding. The reversed position embedding provides the model with duration information on the number of additional tokens to be generated in each step of the inference process \cite{miculicich2023summarization},
and the VAD embedding identifies whether each frame is voice-active or not. Isochrony information is then represented by the sum of the three embedding sequences, which is then concatenated with the semantic feature $S$ before being fed into the decoder through cross-attention. This setup enables our model to align with both the total length and the voice-active regions of the input speech, which is especially beneficial for translating long speeches with pauses.

\subsection{Multi-task Training}\label{sec:multi-task}
We claim that this translation model is a cascaded trainable end-to-end model. Because the distribution $P(Y'|S, A, I, T)$ and $P(T|S, I)$ can be fitted both separately and jointly using various datasets. It significantly enhances data accessibility, which is crucial given the scarcity of S2ST datasets. We utilize these datasets during training as follows:
\begin{itemize}[leftmargin=*]
    \setlength{\topmargin}{0pt}
    \setlength{\itemsep}{0.1em}
    \setlength{\parskip}{0pt}
    \setlength{\parsep}{0pt}
    \item \textbf{S2ST Dataset} comprises quadruple of ($X$, $T_s$, $T_t$, $Y$), where $T_s$ is the source text and $T_t$ is the target text. We either use $X$ or $T_s$ as input and consecutive $T_t$ and $Y$ as target labels, or we can reverse the roles of source input and target labels. 
    \item \textbf{ST Dataset} is a triplet of ($X$, $T_s$, $T_t$). In our model, the missing codec label does not impede the training of speech-to-text translation. We can use $X$ as the input and treat $T_t$ as an incomplete label that terminates with a separation token. In this setup, the isochrony feature is omitted and does not concatenate with the semantic feature. And, alternatively, $T_t$ can serve as the input with consecutive $X$ and $T_s$ as the label, mirroring the approach in the S2ST dataset.
    \item \textbf{ASR Dataset}, which is highly accessible, consists of pairs ($X$, $T_s$). Unlike most other speech translation works, we use ASR data mainly for TTS training, fitting $P(Y'|S, A, I, T)$. Here, $T_s$ is used as the input with consecutive $T_s$ and $X$ as the output labels, and the loss is not applied to the part of the label corresponding to $T_s$. We didn't train on the recognition task because our semantic encoder is frozen, making this task ineffective.
\end{itemize}

\section{Codec and Acoustic Modeling}\label{sec:other}
\subsection{Nerual Codec with Semantic Distillation}
\label{codec_section}
Our neural codec is designed to provide high reconstruction quality and compatibility with language model predictions. We refer to our codec as the {\bf S}emantic-{\bf A}ware {\bf S}peech {\bf Codec} (SASCodec), which integrates two existing methodologies: DAC \cite{kumar2023HighFidelitya} and SpeechTokenizer. DAC employs innovative factorized and L2-normalized codes in vector quantization and finely tuned hyperparameters to achieve state-of-the-art reconstruction quality. SpeechTokenizer, on the other hand, uses a Hubert encoder as a teacher and performs semantic distillation between it and the first-layer Residual Vector Quantization (RVQ) codes. This process enriches the RVQ codes with additional semantic information, facilitating downstream language modeling tasks.

We have adopted SpeechTokenizer as our foundational model and made improvements. Originally trained with limited monolingual data, we have expanded the training to include multilingual datasets. Furthermore, we replaced the Hubert encoder with an MMS\cite{pratap2024Scaling} encoder to enhance multilingual semantic distillation. To further improve the efficacy of semantic distillation, we use a fixed SpeechTokenizer model as the teacher and apply L1 mel loss to the audios reconstructed using only the first quantization layer.

To enhance reconstruction quality, we have integrated the codec with factorized and L2-normalized codes. The discriminator, loss configuration, and training hyperparameters align with those used in DAC.

\subsection{Textless Non-autoregressive Acoustic Modeling}
\label{nar_section}
Similar to previous work, the function of our non-autoregressive model is to enhance the acoustic details of generated speech by predicting the subsequent layers of Residual Vector Quantization (RVQ) codes based on the initial layer. We utilize a standard transformer equipped with adaptive layer normalization for this task. The model inputs the summation of embeddings from the first $n$ layers of RVQ codes to predict the $(n+1)$th layer, where $n = 1, 2, ..., N$ and $N$ represent the total number of RVQ layers. A prompt is concatenated with the input embeddings along the time dimension. The prompt is the summation of full $N$ layers of embeddings of the prompt speech. During the training phase, the prompt and the part to be generated are in the same language. However, in the inference phase, they are in two distinct languages for the translation task. This creates a certain gap. Fortunately, our model trained using a multilingual dataset can effectively generalize and bridge this gap.

Our approach diverges from prior studies by excluding textual or phonetic inputs, rendering it a completely textless model. Previous research indicated that omitting semantic information such as text or phonemes typically significantly decreases intelligibility. However, our experiments demonstrate that the textless model either matches or surpasses the performance of models with text inputs, owing to semantic distillation within the SASCodec. 

The textless model offers two significant advantages: Firstly, it improves data accessibility, as the acoustic model can now be trained entirely with unsupervised data. Unlike supervised data, which requires aligned text labels and precise speech clip boundaries, unsupervised data can be arbitrarily segmented, effectively augmenting the dataset. Secondly, traditional non-autoregressive models that utilize phoneme inputs depend on accurate transcriptions to establish alignment and initiate generation. This requirement can be challenging, especially in zero-shot speech translation tasks where users might not provide transcriptions. Our textless model addresses this problem by eliminating the need for transcription.

\paragraph{Layer Beam Search based on Sampling}
\label{sec:lbs}
In previous work, the decoding of non-autoregressive models is typically performed greedily, as the decoding techniques used in auto-regressive models cannot be directly applied. Greedy decoding often leads to an "early decision error" problem where the model cannot revise previous outputs once generated. To address this, we propose a method called Layer Beam Search (LBS), which adapts the beam search algorithm for non-autoregressive models. 


Consider a beam size of $B$ and a vocabulary size of $V$. In a typical beam search algorithm, a beam maintains $B$ hypotheses. At each decoding step, each hypothesis calculates the probability distribution over $V$, yielding $B \times V$ candidates. These candidates are then ranked by their aggregated scores, and the top $B$ are selected to form the new beam. This approach is impractical in non-autoregressive models due to the enormous vocabulary size at each decoding step. For example, for a speech segment containing $L$ tokens, and a codec codebook size of $C$, the effective vocabulary size becomes $|V| = C^L$, which is too large to enumerate.

To tackle this, we employ a sampling method to choose a reasonable number of candidates. Rather than sampling from the entire vocabulary space $V$, we sample from the top-K candidates at each token, thus creating a reduced vocabulary subspace $V'$ of size $k^L$. We sample $N$ times from $V'$ for each candidate in the original beam to form a total of $B * N$ new candidates. We then rank and form a new beam of size $B$ from these sampled candidates. The pseudo code for the Layer Beam Search is provided in appendix \ref{sec:code}.

\section{Experiments and Results}

\subsection{Experimental Settings}
In this section, we introduce the implementation, training, and evaluation of the TransVIP system. We conducted our experiment in the English-French mutual translation setting, as it had relatively more data than other language pairs.  Appendix \ref{sec:metrics} and \ref{sec:detail} include more details about our models and metrics.

\paragraph{Implementation}
All three models within the system are trained using 32 NVIDIA V100 32G GPUs. We utilize Fairseq2 libriary\footnote{https://github.com/pytorch/fairseq2, MIT License} for model building and the PyTorch Lightning framework\footnote{https://lightning.ai/docs/pytorch/stable/, Apache 2.0 License} for distributed data parallel (DDP) training. The training time for each of the three models is around one week. 

\paragraph{Evaluation}
For speech-to-speech translation translation evaluation, we use a subset of CVSS-T \cite{jia-etal-2022-cvss} fr-en test set containing 300 utterances. It is the first three hundred rows in the original table\footnote{https://dl.fbaipublicfiles.com/covost/covost\_v2.fr\_en.tsv.tar.gz} for test set provided by CoVoST 2 \cite{wang2020CoVoST}. We use source speech as the prompt. All the similarity metrics are also calculated with the source speech as the reference.

We use at most, a 10-second prompt for the joint translation model and a 5-second prompt for the NAR acoustic model. If the source speech is shorter than that limitation we just use the whole utterance as the prompt. For baseline, we compared our model with a textless S2UT model and two Seamless models. The textless S2UT model is the All-to-English XM Transformer\footnote{https://github.com/facebookresearch/fairseq/tree/ust/examples/speech\_matrix} release by SpeechMatrix\cite{duquenne2023SpeechMatrix}. For seamless models, We compared with two versions, the medium version, and the expressive version. The medium version uses the same speech-to-text translation model as the one we use to initialize our joint translation model, while the expressive version has larger model size and a stronger text-to-speech model. For codec evaluation, we employ LibriSpeech \cite{panayotov2015Librispeech} test-clean, which is widely adopted for codec and TTS evaluation.  We compare our SASCodec with Encodec \cite{defossez2023High}, Speechtokenizer \cite{zhang2023SpeechTokenizera} and DAC\cite{kumar2023HighFidelitya}. We evaluate S2ST performance on translation(BLEU), speaker\&prosody similarity(SIM \& AutoPCP), Isochrony control(Rate, Pause \& SLC), and naturalness. Please refer to Appendix \ref{sec:metrics} for the details of evaluation matrics.

\subsection{S2ST Evaluation Result}
In this section, we will show the evaluation result of TransVIP in translation performance, voice preservation, isochrony control, and speech naturalness. 

\begin{table}[ht]\small
\centering
\caption{The subjective evaluation results of our model and baseline model. In all scores,  higher values are better. The results of baseline models are inferred from official checkpoints. We have also conducted a statistical significance test between TransVIP and SeamlessExpressive at a 0.05 significance level. Any results that are statistically significantly better are highlighted in bold. Abbreviation: A.PCP(AutoPCP), Nat.(Natureness)}
\label{tab:main}
\resizebox{\textwidth}{!}{
    \begin{tabularx}{1.04\textwidth}{lc *{9}{Y}}
    \toprule
                        & \multirow{2}{*}{\#Size} & \multicolumn{2}{c}{BLEU} & \multirow{2}{*}{SIM}& \multirow{2}{*}{A.PCP}    & \multirow{2}{*}{Rate}  & \multirow{2}{*}{Pause} & \multicolumn{2}{c}{SLC$_p$} & \multirow{2}{*}{Nat.}  \\ 
                        &       & Speech       & Text      &           &           &       &           & 0.2       & 0.4   &              \\ 
    \midrule
    \\[-1em]
    \multicolumn{11}{c}{Dataset}\\
    Source audio        & -     & -         & -         & -         & -         & -     & -         & -         & -     & 2.62           \\ 
    CVSS-T target       & -     & -         & -         & 0.205     & 2.30      & 0.42      & 0.47  & 0.56      & 0.88  & 3.50       \\ \midrule
    \\[-1em]
    \multicolumn{11}{c}{Fr - En}\\
    \\[-1em]
    ST + StyleTTS       &-      & 33.57     & -         & 0.173     & 2.74      & 0.33      & 0.51  & 0.56      & 0.85  & 3.25   \\
    \\[-0.7em]
    Textless S2UT XM    &1.2B   & 15.45     & -         & 0.035     & 1.96      & -         & -     & 0.52      & 0.79  & 3.18        \\
    SeamlessM4T(M)      &1.0B   & 28.95     & 34.69     & 0.037     & 2.27      & 0.16      & 0.52  & 0.43      & 0.79  & 3.08        \\
    SeamlessExpressive  &1.7B   & 30.85     & 35.26     & 0.256     & 2.74      & 0.22      & 0.51  & 0.50      & 0.79  & 2.91        \\
    TransVIP            &1.1B   & \bf 32.60 & 35.34     & \bf 0.320 & 2.49      &\bf0.55    & 0.44  & \bf 0.70  &\bf0.91&\bf3.19  \\
    \midrule
    \\[-1em]
    \multicolumn{11}{c}{En - Fr}\\
    \\[-1em]
    ST + ValleX         &-      & 22.50     & 34.89     & 0.418     & 2.87      & 0.27      & 0.54  & 0.65      & 0.89  & 3.32   \\
    \\[-0.7em]
    SeamlessM4T(M)      &1.0B   & 21.06     & 32.41     & 0.033     & 2.31      & 0.16      & 0.53  & 0.51      & 0.89  & 3.20       \\
    SeamlessExpressive  &1.7B   & 27.39     & 34.89     & 0.335     & 2.53      & 0.30      & 0.52  & 0.58      & 0.92  & 3.57       \\
    TransVIP            &1.1B   & 27.28     & 33.02     & \bf 0.395 &\bf 2.67   &\bf0.45    &\bf0.65& \bf 0.81  &\bf0.99& 3.40       \\
     \bottomrule
    \end{tabularx}
}
\end{table}

\paragraph{Translation Performance}
We evaluated the translation performance using BLEU and ASR-BLEU metrics as presented in Table \ref{tab:main}. 1) By comparing our models, TransVIP and the Seamless baseline, with the state-of-the-art (SOTA) textless S2UT model and analyzing the generated samples, we observed that both our models significantly outperform the textless S2UT model. Notably, the textless model exhibits severe issues with repetition and "illusion" (generating audio that seems plausible but is unrelated to the input), which have been substantially mitigated in both the Seamless model and our TransVIP. This confirms our hypothesis that including an intermediate text is crucial for successful S2ST generation. 2) Our TransVIP demonstrates highly competitive results compared to the Seamless baselines. In the French-to-English translation direction, our model achieved a BLEU score of 32.60, 3.65 points higher than the SeamlessM4T medium and 1.75 points above the larger SeamlessExpressive. In the English-to-French direction, our model surpasses the SeamlessM4T medium by 6.22 points. Additionally, the ASR-BLEU performance of our model is comparable to that of SeamlessExpressive, with only a slight difference of -0.11 points despite a 1.87-point gap in the text BLEU score. The disparity in text BLEU scores is attributed to differences in model size, and our limited S2TT data precludes rectifying this issue.

\paragraph{Voice Preservation}
We assess speaker similarity in the SIM column of Table \ref{tab:main}. The Textless S2UT model and SeamlessM4T Medium do not effectively preserve speaker identity, as evidenced by their near-zero similarity scores. In contrast, our TransVIP model achieves scores of 0.320 and 0.395 in the fr-en and en-fr directions, respectively, slightly outperforming the SeamlessExpressive baseline, which scores 0.256 and 0.335. TransVIP also exceeds the fr-en target speech from the CVSS-T dataset, used in our training, by 0.115 points. This indicates that our model's similarity performance transcends the limitations of its training dataset. Regarding prosody similarity, TransVIP's AutoPCP score surpasses Textless S2UT and SeamlessM4T Medium. Compared to SeamlessExpressive, TransVIP performs 0.25 points lower in the fr-en direction but 0.14 points higher in en-fr. It is significant to note that, unlike TransVIP, SeamlessExpressive explicitly incorporates a prosody encoder.

\paragraph{Isochrony Control}
We assess isochrony control through several metrics: speech length compliant(SLC), speech rate compliant, and pause number compliant. 1) TransVIP demonstrates superior length control, achieving an improvement of no less than 0.18 in Fr-En and 0.23 in En-Fr for SLC$_{0.2}$ compared to baseline models without a length control strategy. 2) TransVIP achieves the highest speech rate compliance among all baselines, benefiting from an acoustic encoder that incorporates rate information into the model. 3) Regarding pause number, TransVIP scores best in En-Fr and slightly underperforms compared to SeamlessExpressive in Fr-En. Upon reviewing the samples, we attribute this to TransVIP's addition of new pauses in the speech to adjust the total duration.


\paragraph{Speech Naturalness}
We observed that the quality of the conditioning speech prompt significantly influences the naturalness of the generated speech, as measured by the NISQA score. For instance, when processing low-quality real French data (Fr-En), SeamlessExpressive registers the lowest score at 2.91. Conversely, with high-quality synthesized English data (En-Fr) as input, SeamlessExpressive achieves the highest score of 3.57. Meanwhile, our TransVIP model demonstrates consistent performance, scoring 3.19 in the Fr-En direction and 3.40 in the en-fr direction. 

\subsection{Codec Evaluation Result}
We show the evaluation result in Table \ref{tab:codec_resyn} and group the result by bandwidth(BW). The SpeechTokenizer that has undergone semantic distillation performs well in the NISQA score, producing a clean and natural voice. On the other hand, DAC performs well in similarity and WER, showing a strong capability of a faithful reconstruction of the original speech. Finally, our SASCodec successfully combines the strength of two codecs, showing a very competitive result in all four metrics.

\begin{table}[ht]\small
\centering
\caption{Codec re-synthesize evaluation result. The codecs are grouped by bandwidth. * means the result is inferred from the official checkpoint. \dag means it is reproduced result using the official recipe. Abbreviation: NQ(number of quantizers), FR(frame rate), BW(Bandwidth), Nat.(Natureness)}
\label{tab:codec_resyn}
\begin{tabular}{@{}lccccccc@{}}
\toprule
            & NQ& FR    & BW  & SIM$\uparrow$   & SIM-o$\uparrow$ & Nat.$\uparrow$ &WER$\downarrow$ \\ 
\midrule
SpeechTokenizer*       & 8& 50    & 4 kbps & 0.847       & 0.558     &\bf 3.714 & 1.44\\
SASCodec       & 8 & 50    & 4 kbps  & \bf 0.871       & \bf 0.576     &  3.425 & \bf 1.34\\
\midrule
Encodec*    & 8 & 75    & 6 kbps  & 0.887       & 0.582     & 3.139 & 1.30\\
DAC*        & 8 & 75    & 6 kbps  & 0.900       & 0.590     & 3.387 & 1.31\\
\midrule
DAC\dag   & 16& 50    & 8 kbps  & 0.936       & 0.622     & 3.667 & 1.11\\
SASCodec       & 16& 50    & 8 kbps  &\bf 0.939    &\bf 0.624  &\bf 3.785 & 1.11\\
\midrule
Encodec*    & 16& 75    & 12 kbps & 0.931       & 0.618     & 3.281 & 1.16\\
DAC*        & 16& 75    & 12 kbps &\bf 0.967    &\bf 0.646  & 3.636 & 1.07\\
SASCodec       & 24& 50    & 12 kbps & 0.950       & 0.631     &\bf 3.786 & \bf1.06\\

\bottomrule
\end{tabular}
\end{table}

\section{Ablation Study}\label{sec:ablation}

\subsection{Ablation on Acoustic Embedding}
In the process of semantic distillation within the codec, a significant amount of acoustic information was disentangled from the first-layer codec. This raises the question of whether the acoustic encoder is still necessary in the joint translation model. To investigate this, we leveraged the model's capability to perform inference without the acoustic embedding. We conducted such inferences and compared the outcomes, as shown in Table \ref{tab:abl_acu}. Without the acoustic embedding, speaker similarity scores decreased by 0.040 in the French-English (Fr-En) translations and by 0.032 in the English-French (En-Fr) translations. Additionally, there was a slight decline in the AutoPCP scores. These findings suggest that residual acoustic information remains within the first-layer codec, underscoring the necessity of including a prompt embedding.

\begin{table}[ht]\small
\centering
\caption{The ablation study of the acoustic embedding in the joint translation model.}
\label{tab:abl_acu}

\begin{tabularx}{0.62\textwidth}{lc *{5}{Y}}
    \toprule
                        &ASR-BLEU& BLEU &SIM & A.PCP  & Nat. \\ 

    \midrule
    \\[-1em]
    \multicolumn{6}{c}{Fr - En}\\
    \\[-1em]
    TransVIP            &  32.60    & 35.34     &  0.320    & 2.49      &3.19       \\
    \quad  -A\_emb    &  32.47    & 35.18     &  0.280    & 2.45      &3.23       \\ \midrule
    \\[-1em]
    \multicolumn{6}{c}{En - Fr}\\
    \\[-1em]
    TransVIP            & 27.28     & 33.02     &  0.395    & 2.67      & 3.40      \\
     \quad  -A\_emb   & 26.84     & 33.15     &  0.362    & 2.45      & 3.46      \\ \bottomrule
\end{tabularx}
\end{table}

\subsection{Ablation on Isochrony Control method}
We compared our proposed Isochrony control method with and without using other strategies. We presented the results in Table \ref{tab:IC}, where a) No Isochrony Control (No IC). b) Isochrony Control on the Decoder (Dec IC). This involves adding the Isochrony embedding to the input of the encoder as another positional embedding. We implemented the method from \cite{wu2022VideoDubber} in our system. c) Isochrony Control on the Decoder with Future Pause Information (Dec IC + FPI). This is an improvement over (b). In addition to the distance to the global end and VAD information, two extra pieces of information are encoded: the distance to the next pause and the number of pauses in the future. We implemented the method from \cite{pal2023improving} in our system.

It demonstrates that our proposed method reaches the best isochronic control. The results also show that this approach can improve the ASR-BLEU score compared to Isochrony control on the decoder, meaning the model is more confident and accurate in the generation and makes fewer errors like repetition and truncation.

\begin{table}[htbp]\small
\centering
\caption{Ablation study on the isochrony control strategy.}
\begin{tabularx}{0.7\textwidth}{lc *{4}{Y}}
\toprule
      &ASR-BLEU& Overlap&   SLC$_{0.2}$ &SLC$_{0.4}$   \\
\midrule
No IC       & 30.81         & 0.689      & 0.63      & 0.87        \\
Dec IC      & 30.51         & 0.748      & 0.75      & 0.90           \\
Dec IC+ FPI & 30.45         & 0.766      & 0.77      & 0.91       \\
Enc IC (Proposed)      & 30.62         & 0.784      & 0.82      & 0.95        \\
\bottomrule
\end{tabularx}
\label{tab:IC}
\end{table}

\subsection{Ablation Study on the NAR Acoustic Model}\label{sec:abl_nar}

In this section, we conduct two comparisons. Firstly, we contrast the performance of a textless non-autoregressive acoustic model with that of a model trained using text transcriptions (BPE) as input. Secondly, we assess the inference results with and without the utilization of the Layer Beam Search (LBS) algorithm to determine its impact on performance enhancement.

The textless model is trained on a combination of datasets including Librilight, VoxPopuli French subset, SeamlessAlign, and Common Voice English and French subsets. As some datasets do not provide text transcriptions, we substitute the LibriHeavy dataset in the model trained with text input. Additionally, in the textless model, long audio segments are randomly clipped, whereas in the model with text input, clips strictly adhere to timestamps provided in the metadata to align with text transcriptions. We compare the Fr-En uni-directional setup to mitigate the impact of the absence of a large-scale French dataset on the model with text input. Despite the apparent data discrepancy, textless modeling demonstrates superior performance, underscoring one of its key data accessibility advantages. All other hyperparameters, including training steps, remain consistent across both models.

The results are presented in Table \ref{tab:nar}. The textless model consistently outperforms the model with text input across all metrics of ASR-BLEU, speaker similarity, and naturalness. Moreover, employing LBS yields superior results compared to greedy decoding.

\begin{table}[htbp]\small
\centering
\caption{Ablation study on the NAR Acoustic model and its inference strategy. }
\begin{tabular}{lccc}
\toprule
NAR Model & SIM & ASR-BLEU & Nat.  \\
\midrule
NAR w/o text            & 0.320 & 32.60 & 3.19 \\
\quad - LBS               & 0.309 & 32.30  & 3.17 \\
\midrule
NAR w/ text   & 0.307 & 31.52  & 3.10 \\
\quad - LBS      & 0.298 & 31.03  & 3.09 \\
\bottomrule
\end{tabular}
\label{tab:nar}
\end{table}

\section{Conclusion, Limitations and future works}
In this paper, we develop a speech-to-speech translation framework that preserves speaker voice and isochrony, generating high-quality translated speech suitable for both daily communication and automatic video dubbing. Our framework can infer in an end-to-end manner, jointly considering text and speech probability during the inference phase and making full use of various data during the training phase. In the Fr-En direction, TransVIP surpasses the current state-of-the-art model with comparable translation performance and significantly improved voice and isochrony preservation. 

\textbf{Limitations and future works:} 
1) Performance \& Language Support: Due to current resource and data limitations, we have only trained our model on the Fr-En language pair. We used approximately 5k hours of audio to train the translation model, compared to over 50k hours in previous works\cite{communication2023SeamlessM4T, wang2023VioLA, dong2023PolyVoice}. This suggests that there is potential for performance improvement if we scale up the data. Additionally, we aim to extend our framework to many-to-many settings through large-scale multilingual training, similar to Seamless. 2) More Detailed Attribute Control: Upon reviewing the test samples, we found that TransVIP occasionally alters the tone of the speech, such as translating interrogative sentences into declarative ones. It is reasonable to hypothesize that more detailed control of attributes like intonation or emotion could help address this issue.

\textbf{Broader Impact:} TransVIP was developed to improve cross-lingual communication and assist with automated video dubbing workflows. However, it carries potential risks, including the misuse of the model for impersonating specific speakers by inputting targeted text and speaker prompts. Additionally, as a generative model, there is a possibility it may produce toxic or biased outputs, although such occurrences were not observed in our test samples.

\begin{ack}
We extend our gratitude to Chung-Hsien Tsai, Canrun Li, and Zhen Xiao for their invaluable contribution in conducting the subjective evaluation and for their efforts in assembling the video dubbing material. 
\end{ack}

\medskip

\small
\bibliographystyle{IEEEtran}
\bibliography{r}

\newpage
\appendix

\section{Evaluation Metrics}\label{sec:metrics}
For S2ST evaluation, we evaluate translation performance (BLEU and ASR-BLEU), speaker similarity (SIM), prosody similarity(AutoPCP), isochrony control (Rate, Pause, and SLC-0.2/0.4), and naturalness (NISQA). The score is the higher the better for all the S2ST metrics. 
\begin{itemize}[leftmargin=*]
    \item  BLEU is calculated between text output and text label using the 'corpus\_bleu' method of sacrebleu\footnote{https://github.com/mjpost/sacrebleu, Apache 2.0 License.}\cite{post-2018-call}. To calculate speech BLEU(also known as ASR-BLEU), we utilize a whisper large v3 model to transcribe the generated speech into text and then calculate the BLEU score between the transcribed text and text label.
    \item The Speaker Similarity(SIM) score is obtained by calculating the cosine distance of the speaker embedding of two sentences. The embedding comes from a WavLM\cite{chen2022WavLMa}-based speaker verification model\footnote{https://github.com/microsoft/UniSpeech/tree/main/downstreams/speaker\_verification, CC BY-SA 3.0 Unported License}. In the context of S2ST evaluation, this SIM score is generally computed by using two embeddings extracted from source speech and translated target speech, respectively. This SIM score reported is typically lower than that of zero-shot TTS since the two embeddings used for the computation are extracted from speech in different languages. 
    \item AutoPCP\footnotemark[\getrefnumber{fn:seamless}] is a model-based evaluation tool for Prosodic Consistency Protocol(PCP)\cite{huang2023Holistic} which evaluate overall prodsodic similarity.
    \item Rate\footnote{\label{fn:seamless}https://github.com/facebookresearch/seamless\_communication/blob/main/docs/expressive/README.md, CC-BY-NC 4.0 License} calculated correlation in speaking rate (measured by syllables per second). Pause\footnotemark[\getrefnumber{fn:seamless}] calculate the correlation in pause number. Text transcription is needed for the rate and pause calculation. Therefore they are not suitable for textless S2ST models.
    \item SLC$_p$(Speech Length Compliant) is proposed in \cite{wu2022VideoDubber} that calculates the fraction of generated speech with duration within $[1-p, 1+p]$ times of the original speech input.
    \item 
    For naturalness evaluation, we use NISQA-TTS(V1.0)\footnote{https://github.com/gabrielmittag/NISQA, MIT License}, which is a model to evaluate the naturalness and quality of synthesized speech. \cite{mittag2020Deep}
\end{itemize}

For codec evaluation, we use metrics that are commonly employed in zero shot-TTS to evaluate the re-synthesize quality of the codec, which are 1) SIM, 2) NISQA, and 3) WER. The NISQA metrics are the same as those in the above S2ST evaluation. A market-leading Speech Recognition API is used for Word Error Rate (WER) calculation. It measures the correctness and intelligibility of re-synthesized speech. We use two similarity scores. SIM is used to calculate the similarity between original speech and resynthesised speech. Generally, this SIM value is very high because the two embeddings used in the calculation are extracted from speech that shares the same content in the same language. Following \cite{wang2023neural, le2023Voicebox}, the SIM-o metric is computed by first segmenting the audio into two parts: a 3-second prompt and the remaining portion. The latter part is then re-synthesized. Finally, similairy score is calculated between the re-synthesized audio and the original prompt.

\section{Implementaion Details}\label{sec:detail}

\paragraph{Joint Translation model} The primary encoder-decoder (Enc-Dec) architecture of the joint model is initialized using the SeamlessM4T S2T model, while the other submodules are trained from scratch. The first-layer RVQ codes are integrated into the vocabulary of the foundational S2T model. As mentioned in Section \ref{sec:multi-task}, this model is trained using multiple datasets, including two S2ST datasets: CVSS-T\cite{jia2022CVSS} and SeamlessAlign\footnote{https://github.com/facebookresearch/seamless\_communication/blob/main/docs/m4t/seamless\_align\_README.md}, one internal ST dataset, and one ASR dataset: Common Voice\cite{ardila2020Common} version 15 (English and French subsets). An offline preprocessing step is performed to convert the speech labels into a discrete codec form. During this process, we utilize a VAD tool\footnote{https://github.com/snakers4/silero-vad, MIT License} to eliminate silence at the beginning and end of the speech recordings. This VAD tool is also used to generate the isochrony embeddings during both the training and inference phases. The input to the speech semantic encoder is Fbank features and text is tokenized into BPE, which is consistent with the Seamless framework. The acoustic encoder is a six-layer standard Transformer encoder with a hidden size of 1024. Its input features are extracted from raw waveforms by the SASCodec encoder, providing richer acoustic information than Fbank features. For decoding we employed a beam search algorithm with a beam size set to 5.
\paragraph{NAR Acoustic Model} This 12-layer transformer model is trained from scratch and utilizes mainly two unsupervised corpora: LibriLight\cite{kahn2020LibriLight}, containing 52k hours of English speech, and VoxPopuli\cite{wang2021VoxPopuli} French subset, encompassing 23k hours of French speech. We also included audio from SeamlessAlign and Common Voice in the training. The datasets are weighted differently during training to ensure that data from both languages are sampled with equal probability. In inference, we employ the LBS mentioned in section \ref{sec:lbs}. There are 16 RVQ layers in the model, which means 15 generation steps. The beam size is 10, the sampling number is 20 and $K$ is 3.
\paragraph{SASCodec Model} We train the model on audio with a frequency of 16k Hz. The downsample rate is set at 320, resulting in 50 tokens/frames per second, instead of the 75 tokens used in the official DAC checkpoint. This adjustment is primarily made to ensure a consistent token rate that matches the semantic token rate, thereby facilitating the distillation process.
The model is trained with 24 RVQ codebooks while only 16 are employed in the translation system. The SASCodec model is trained using the full set of the Common Voice version 15 corpus, which includes hours of multilingual speech data. During each training iteration, a random 2-second segment is cropped from the speech samples.

\section{Additional Ablation on the Choice of Codec}
In this section, we compare training TransVIP using different codecs: SpeechTokenizer \cite{zhang2023SpeechTokenizera} and DAC \cite{kumar2023HighFidelitya}. In this ablation study, the joint translation model is trained with a subset containing only CVSS-T Fr-En uni-direction data. For the NAR acoustic model, both DAC and SASCodec use 16 codec layers, while SpeechTokenizer uses 8 layers, as it only has an 8-layer version. The results are shown in Table \ref{tab:codec}. Compared to SpeechTokenizer, the model trained with SASCodec exhibits superior performance in all aspects. Most notably, the speaker similarity improved by 0.04, from 0.226 to 0.264, aligning with the improvement in codec re-synthesis results.

\begin{figure}
    \centering
    \includegraphics[width=\linewidth]{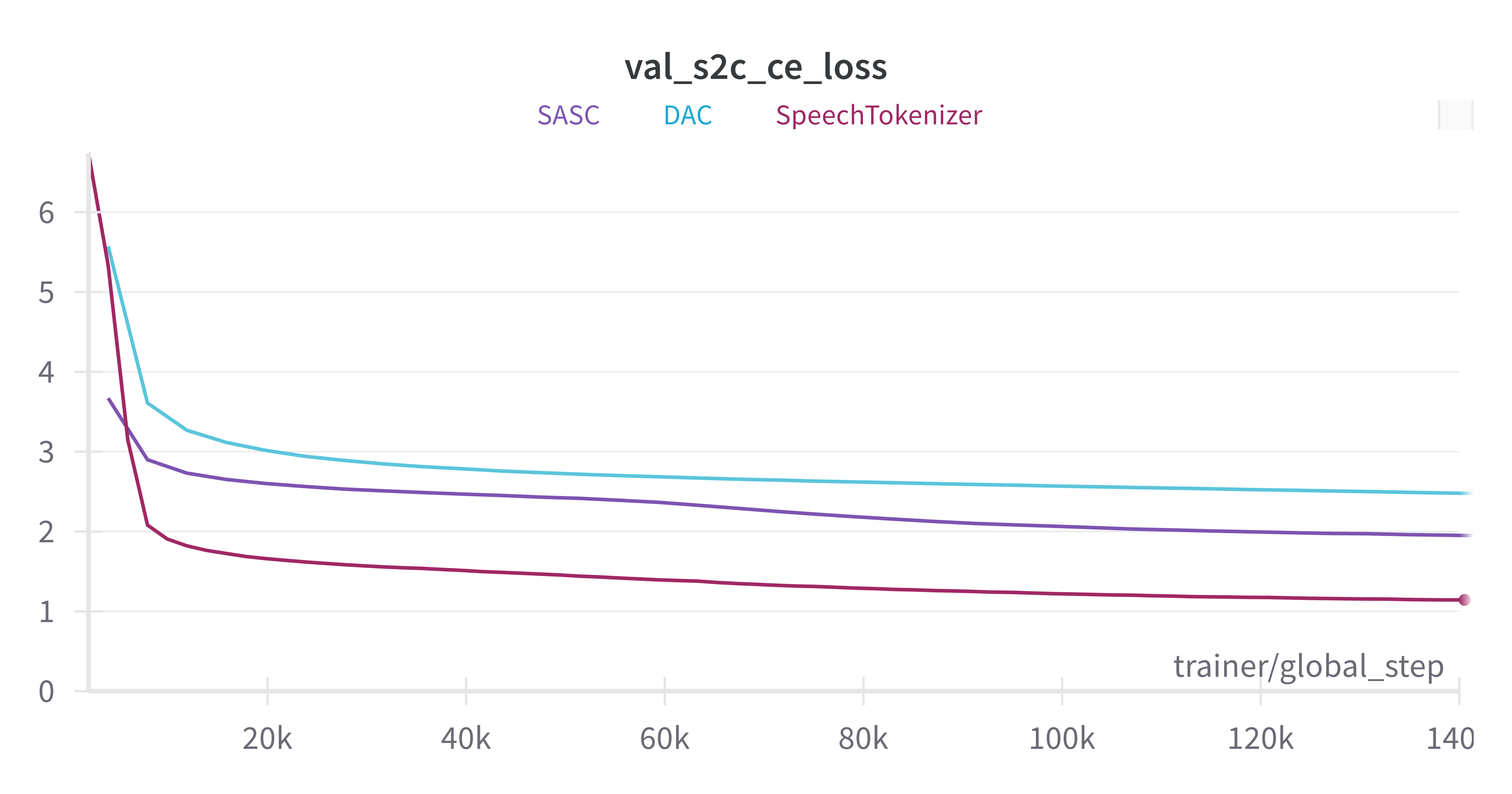}
    \caption{Validation loss of TransVIP using different codecs.}
    \label{fig:loss}
\end{figure}

Additionally, we find that when trained with DAC, the joint translation model fails to generate reasonable output even when the training loss appears to converge. To investigate this, we plot the validation loss curve of the joint translation model in Figure \ref{fig:loss}. It shows that SpeechTokenizer has the lowest loss, followed by SASCodec, and then DAC. Furthermore, we observed that SpeechTokenizer can generate reasonable output with minimal training steps. We attribute this finding to two reasons: 1) SpeechTokenizer contains most semantic information in its first-layer codec. Although SASCodec also undergoes semantic distillation, it still contains acoustic information in the first layer, trading for better speech reconstruction quality. 2) The enhancement of RVQ in DAC, which is also utilized in SASCodec, leads to an even utilization rate across the codebook. This could pose a challenge for language modeling. However, our SASCodec method effectively balances the quality of reconstruction and the affinity of the language model.

\begin{table}[htbp]
\centering
\caption{Ablation study on the choice of codec.}
\begin{tabularx}{0.8\textwidth}{lc *{6}{Y}}
\toprule
 \multirow{2}{*}{Codec Model}     &\multicolumn{2}{c}{BLEU}& \multirow{2}{*}{SIM}&   \multicolumn{2}{c}{SLC$_p$} & \multirow{2}{*}{Nat.}      \\
                & Speech       & Text      &          & 0.2  &0.4  &       \\
\midrule
SpeechTokenizer & 29.81     & 34.18     & 0.226      & 0.76      & 0.93     & 3.02      \\
SASCodec            & 30.62     & 34.30     & 0.264      & 0.82      & 0.95     & 3.09      \\
\bottomrule
\end{tabularx}
\label{tab:codec}
\end{table}

\section{Pseudo code for Layer Beam Search}\label{sec:code}
The pseudo code in the Pytorch style is shown in algorithm \ref{alg:1}.

\begin{algorithm}[H]
\caption{Layer Beam Search}
\label{alg:1}
\DontPrintSemicolon 
\SetKwFunction{Mean}{mean}
\SetKwFunction{Gather}{gather}
\SetKwFunction{Forward}{forward}
\SetKwFunction{LogSoftmax}{log\_softmax}
\SetKwFunction{Softmax}{softmax}
\SetKwFunction{Sample}{sample}
\SetKwFunction{TopK}{top\_k}
\SetKwFunction{Zeros}{zeros}
\SetKwFunction{Zip}{zip}
\SetKwFunction{Cat}{cat}
\SetKwFunction{Append}{append}
\SetKwFunction{Flatten}{flatten}
\SetKwFunction{Expand}{expand}
\SetKwInput{KwData}{Input}
\SetKwInput{KwResult}{Output}
\SetKwFunction{FMain}{GenerateNar}

\KwData{\;
$inputs$: Inputs to the NAR model\;
$n\_codebook$: Number of RVQ layers. Default: 16\;
$beam\_size$: Beam\_size. Default: 10\;
$n\_sample$: Number of sampled candidates of each hypothesis. Default: 20\;
$K$: At each token only sample from the top K candidates. Default: 3\;
}
\KwResult{\;
The same data class as the $inputs$ and contains generation results.\;
}

\SetKwProg{Fn}{Function}{:}{}
\Fn{\FMain{$inputs, n\_codebook, beam\_size, n\_sample, K$}}{
    \For{$\text{layer} \gets 0$ \KwTo $n\_codebook - 1$}{
        $logits \gets \Forward(inputs)$\;
        $lprobs \gets \LogSoftmax(logits, \text{dim}=-1)$\;
        $topk\_values, topk\_ids \gets \TopK(logits, K)$\;
        $candidates \leftarrow \Zeros(beam\_size, n\_sample, seq\_len)$\;
        \For{$i \gets 0$ \KwTo $n\_sample - 1$}{
            $samples \gets \Sample(\Softmax(topk\_values))$\;
            $sampled\_ids \gets \Gather(topk\_ids, samples)$\;
            $sampled\_lprobs \gets \Gather(lprobs, sampled\_ids)$\;
            $avg\_score \gets \Mean(sampled\_lprobs)$\;
            $candidates[:, i] \gets sampled\_ids$\;
            $total\_score[:, i] \gets total\_score[:, i] + avg\_score$\;
        }
        $values, flat\_ids \leftarrow \TopK(\Flatten(total\_score), beam\_size)$\;
        $rows \leftarrow flat\_ids\text{ // } beam\_size$\;
        $cols \leftarrow flat\_ids \text{ \% }beam\_size$\;
        $total\_score \leftarrow values.\Expand(-1, beam\_size)$\;
        $next\_inputs \leftarrow []$\;
        \For{$(row, col)$ \textbf{in} $\Zip(rows, cols)$}{
            $a \leftarrow inputs[row]$\;
            $h \leftarrow candidates[row, col]$\;
            $next\_inputs\text{.\Append}(\Cat([a, h]))$\;
        }
        $inputs \leftarrow next\_inputs$\;
    }
    \KwRet $inputs$\;
}
\end{algorithm}

\section{Model size}
We report in Table \ref{tab:module-size} the detailed decomposition of module sizes of TransVIP and baseline models.
\begin{table}[htbp]
\centering
\caption{Parameters of the building components. Only the modules that are required in the inference stage are calculated.}
\begin{tabular}{@{}lccccc@{}}
\toprule
 & Speech Encoder & Decoder & T2U & NAR & Total \\ 
\midrule
Textless S2UT XM    & 970M & 204M & -    & - & 1174M \\ 
SeamlessM4T-Medium  & 366M & 415M & 170M & - & 951M \\ 
Seamless-Expressive & 635M & 867M & 204M & - & 1706M \\ 
TransVIP            & 445M & 415M & - &244M & 1104M \\ 
\bottomrule
\end{tabular}
\label{tab:module-size}
\end{table}

\section{Data Preprocessing and Pseudo-labeling}
As mentioned in section \ref{sec:consecutive}, text labels play a crucial role in joint translation model training. However, text labels are absent in some end-to-end S2ST datasets such as SeamlessAlign and SpeechMatrix. To utilize these datasets, pseudo text labels are required. We employ the following procedure to preprocess such datasets:
\begin{enumerate}[leftmargin=*]
\item Generate pseudo transcription labels for the source and target speech using an ASR model. In our experiments, we use the Whisper large v3 model.
\item Extract codec from the source and target speech, removing silence at the beginning and end using a VAD model.
\item Validate and score the data using a pre-trained text-to-text translation model. In our experiments, we use the pre-trained mBART-large-50-mmt model.\footnote{https://huggingface.co/facebook/mbart-large-50-many-to-many-mmt} The mean confidence scores in both directions are averaged to determine the confidence score of the data pair.
\end{enumerate}
During training, we filter out data with low confidence scores to avoid potential bad data or ASR errors.

\section{Subjective Evaluation}
\label{CMOS_instruction}
\begin{figure}
  \centering
  \includegraphics[width=1\linewidth]{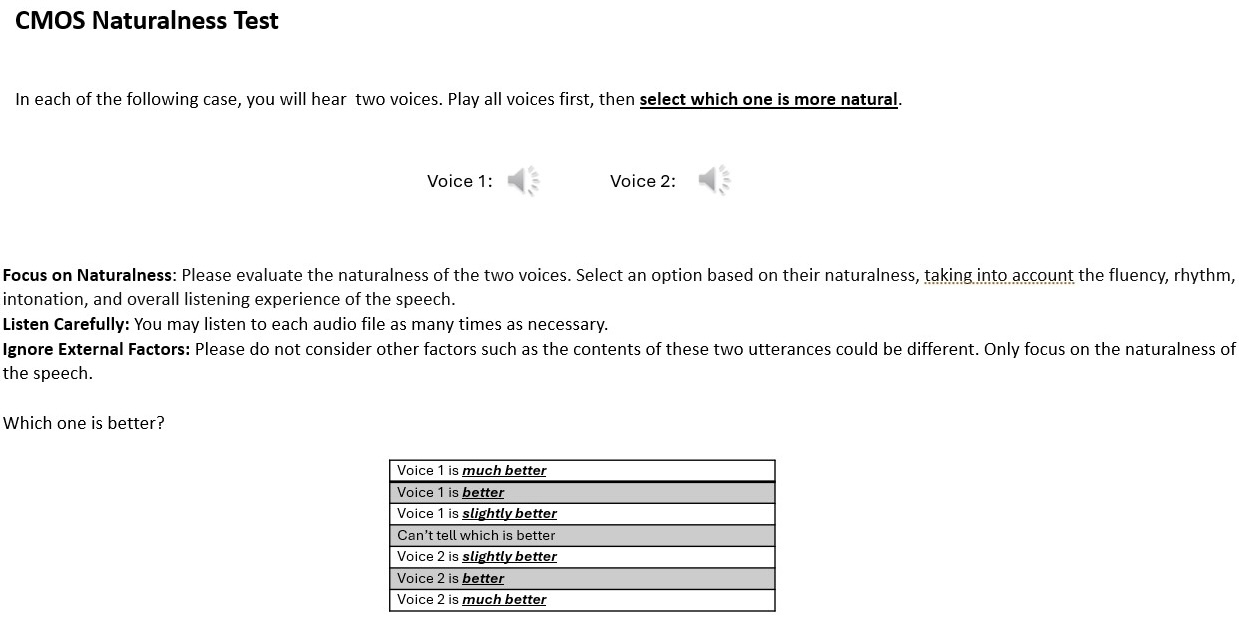}
  \caption{The instruction of CMOS naturalness test}
  \label{fig:cmos}
\end{figure}

Given that the NISQA score can present difficulties in evaluating the quality or naturalness of noisy speech, we also assess the naturalness of speech using a subjective metric known as the Comparative Mean Opinion Score (CMOS). For this evaluation, 12 professional linguistic experts judged 120 pairs of utterances translated by TransVIP and SeamlessExpressive. The results indicated that TransVIP's performance is significantly better (p<0.01) than SeamlessExpressive, i.e., TransVIP outperforms SeamlessExpressive by achieving a preference score higher by 0.46. The instruction of CMOS naturalness test is shown in Figure \ref{fig:cmos}. Much better/better/slightly better/can't tell equals to 3/2/1/0 score respectively.

\end{document}